# BEHAVIOR-SPECIFIC FILTERING FOR ENHANCED PIG BEHAVIOR CLASSIFICATION IN PRECISION LIVESTOCK FARMING

Zhen Zhang[1], Dong Sam Ha[1], Gota Morota [2,3], and Sook Shin[1*]

[1] The Bradley Department of Electrical and Computer Engineering, Virginia Tech, Blacksburg, Virginia,
[2] Department of Animal and Poultry Sciences, Virginia Tech, Blacksburg, Virginia, USA
[3] Laboratory of Biometry and Bioinformatics, Department of Agricultural Environmental
Biology, Graduate School of Agricultural and Life Sciences, The University of Tokyo, Tokyo, Japan

## ABSTRACT

*This study proposes a behavior-specific filtering method to improve behavior classification accuracy in Precision Livestock Farming. While traditional filtering methods, such as wavelet denoising, achieved an accuracy of 91.58%, they apply uniform processing to all behaviors. In contrast, the proposed behaviorspecific filtering method combines Wavelet Denoising with a Low Pass Filter, tailored to active and inactive pig behaviors, and achieved a peak accuracy of 94.73%. These results highlight the effectiveness of behavior-specific filtering in enhancing animal behavior monitoring, supporting better health management and farm efficiency.*

## 1. INTRODUCTION

### 1.1. Background

Precision Livestock Farming (PLF) has emerged as a critical field for monitoring and improving animal health and behavior[1]. Accurate and continuous tracking of livestock behavior is essential for identifying early signs of health issues and enabling timely intervention. Traditional methods for monitoring pig behavior, such as manual observation, are labor-intensive, limited in scalability, and prone to inaccuracies [2].

Recent advancements in PLF have introduced automated systems that leverage biosensors to track behavior in real time. These sensors, often attached to animals, collect data that is both costeffective and reliable, making them indispensable for modern livestock management [3,4]. For example, wearable wireless biosensors have been successfully applied to monitor various physiological parameters and behaviors in livestock, contributing significantly to effective group management [3]. Although much of the initial work has focused on cattle [4,5], there is growing interest in applying similar technologies to pigs. However, despite the advantages of biosensors in enhancing data quality and quantity, accurately processing these data to interpret diverse pig behaviors remains a challenge [6].

### 1.2. Challenges in Behavior Monitoring

Conventional signal processing methods, such as low-pass filters and wavelet denoising, are often employed to reduce noise in sensor data [7-9]. Although these techniques enhance data quality, they typically apply a uniform filtering strategy regardless of the type of behavior observed [10]. This one-size-fits-all approach fails to account for the distinct characteristics of behaviors—active behaviors (e.g., walking, eating) involve rapid, dynamic movements requiring preservation of high-frequency components, whereas inactive behaviors (e.g., lying, standing) benefit from aggressive noise reduction to eliminate spurious signals [11,12]. As a result, applying uniform filtering compromises data accuracy and diminishes the performance of downstream machine learning models for behavior classification and tracking.

### 1.3. Related Works

Recent studies have utilized both accelerometer and gyroscope data combined with machine learning algorithms to classify livestock behaviors more accurately [13-16]. For example, multiaxis sensor data has been employed to distinguish behaviors such as eating, walking, and standing in cattle, resulting in enhanced classification performance [16]. Additionally, deep learning techniques, such as convolutional neural networks (CNNs) and recurrent neural networks (RNNs), have been applied to automatically recognize animal behaviors, leveraging the comprehensive motion information from multi-sensor data [17,18]. However, many approaches do not tailor preprocessing techniques to the specific characteristics of different behavior types, leading to suboptimal performance. This highlights the need for behavior-specific filtering methods, which the current study aims to address.

### 1.4. Proposed Solution and Objectives

To address the limitations of traditional one-size-fits-all filtering methods, this study proposes a tailored behavior-specific filtering approach. This method optimizes sensor data preprocessing by applying distinct filtering techniques based on the specific characteristics of active and inactive behaviors. For example, high-frequency components critical for active behaviors (e.g., walking, eating) are preserved, while noise is reduced for inactive behaviors (e.g., lying, standing).

In this approach, pig behaviors were manually categorized into active and inactive groups, and separate filtering pipelines were applied accordingly. The objective was to compare the performance of this behavior-specific filtering method with that of conventional uniform filtering approaches using standard evaluation metrics, such as accuracy and F1-score. This study aims to establish a foundation for more refined preprocessing strategies in PLF systems, ultimately supporting more reliable behavior monitoring.

## 2. METHODS

### 2.1. Data Collection and Labeling

Animal handling and media recording were approved and conducted in accordance with the Virginia Tech Institutional Animal Care and Use Committee. Data were collected from two pigs at the Virginia Tech Swine Facility over 10 non-consecutive days between October 24, 2022, and November 13, 2022. As shown in Figure 1, each pig was fitted with a MetaMotionC (MMC) sensor attached via standard ear tags, which recorded 6-axis motion data (accelerometer and gyroscope) at 50 Hz, resulting in 2,276,450 data points. An RGB camera mounted on the pen ceiling recorded video at 30 frames per second, and the footage was synchronized with the sensor data. Ground truth labels were manually annotated using SegIt software [19] and subsequently verified, categorizing behaviors into seven classes: Eating, Lying, Walking, Standing, Interacting, Drinking, and Unknown. This synchronized and manually labeled dataset served as the foundation for evaluating the filtering approaches and subsequent modeling.

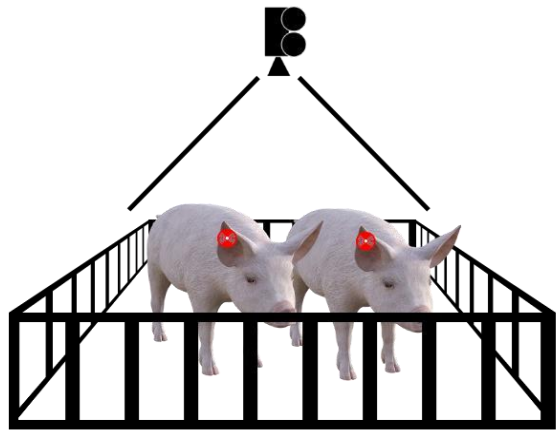

Figure 1. Demonstration of data collection setup: A RGB camera mounted on the ceiling, pigs are equipped with MMC sensors on the ear.

### 2.2. Data Pre-processing

#### 2.2.1. Outlier Detection and Data Imputation

Outlier detection was performed using both Interquartile Range (IQR) and Hampel filters. The Hampel filter was chosen for its ability to effectively identify anomalies with minimal data loss. Following outlier removal, linear interpolation was applied to impute missing values, while preserving the temporal structure of the data.

#### 2.2.2. Filtering Methods

As a baseline, traditional filtering techniques, including Wavelet Denoising, Total Variation Denoising (TVD), Median Filtering, High-Pass Filtering (HPF), Low-Pass Filtering (LPF), and Savitzky-Golay Filtering, were uniformly applied to the sensor data. The proposed method was implemented to account for the distinct signal characteristics associated with different pig behaviors. The behaviors were manually categorized into two groups: *active* (Eating, Walking, Interacting, Drinking) and *inactive* (Lying, Standing). For active behaviors, filtering methods that preserve high-frequency components (e.g., Wavelet Denoising and TVD) were employed to retain relevant motion details, while for inactive behaviors, techniques that reduce noise (e.g., LPF and Median Filtering) were applied to enhance signal stability. This targeted strategy was designed to preserve features that are critical for accurate behavior classification.

## 2.3. Feature Extraction and Selection

The pre-processed sensor data were segmented into 1.5-second windows, following established practices [20,21]. A total of 104 features were extracted from each window to capture various aspects of the signal, including its statistical, spectral, and overall characteristics. These features are grouped as follows:

- **Time-Domain Features**: Statistical metrics such as minimum, maximum, mean, median, variance, percentiles, root mean square (RMS), skewness, and kurtosis were computed for each of the six channels (X, Y, Z for both accelerometer and gyroscope), resulting in 60 features.
- **Frequency-Domain Features**: Spectral metrics including spectral entropy, frequency centroid, energy, and other FFT-derived measures were calculated for each channel, yielding 30 features.
- **Aggregate Features**: Overall signal properties, such as Signal Magnitude Area and summed values per channel and per category, contributed an additional 14 features.

To reduce dimensionality and improve computational efficiency, Recursive Feature Elimination was applied to reduce the feature set to 50 key features. Following feature extraction and selection, all features were normalized to the [0, 1] range using the MinMaxScaler. This normalization ensures consistent scaling across features, which is essential for optimizing model training and evaluation.

## 2.4. Classification Models and Performance Metrics

A diverse set of machine learning models was evaluated to objectively assess the impact of filtering approaches on behavior classification. The preprocessed dataset was randomly split into 70% for training and 30% for testing. This split ensured that model performance was evaluated on unseen data, providing a reliable measure of generalization.

The models used in this study were selected to represent various classification strategies and capture different algorithmic strengths. These models include:

- **Random Forest (RF)**: An ensemble method that aggregates predictions from multiple decision trees, effectively handling high-dimensional data and reducing overfitting.
- **XGBoost (XGB)**: An optimized gradient boosting algorithm designed for efficiency and scalability, capable of capturing complex feature interactions.
- **K-Nearest Neighbors (KNN)**: A non-parametric method that classifies instances based on proximity in the feature space, albeit with higher computational cost for large datasets.
- **Gradient Boosting Machine (GBM)**: An iterative ensemble technique that builds decision trees sequentially, minimizing classification errors at each step.
- **Decision Tree (DT)**: A simple, interpretable model that recursively splits data based on feature values, although it can be prone to overfitting when used alone.
- **Linear Support Vector Classifier (Linear SVC)**: A linear variant of support vector machines that is effective in high-dimensional spaces, provided the classes are approximately linearly separable.
- **Naive Bayes (NB)**: A probabilistic classifier based on Bayes' theorem, offering computational efficiency while assuming feature independence.

Each model was trained on the training set and evaluated on the testing set. Performance was measured using the following metrics:

- **Accuracy**: Measures the overall proportion of correctly classified instances.
- **Precision**: Indicates the proportion of positive identifications that were actually correct.
- **Recall**: Reflects the proportion of actual positives that were correctly identified.
- **F1-Score**: The harmonic mean of precision and recall, providing a balanced assessment of model performance, particularly in cases of class imbalance.

This comprehensive evaluation facilitates a comparative analysis between traditional uniform filtering methods and the proposed behavior-specific filtering approach, highlighting the tradeoffs and benefits of each technique.

### 2.5. Cross-Validation Evaluation

In addition to hold-out validation, a 10-fold stratified cross-validation was performed on the dataset. The data was split into 10 folds, each maintaining the original class distribution. In each iteration, the model was trained on 9 folds and tested on the remaining fold. Accuracy, precision, recall, and F1-score were calculated for each fold, and the final results are reported as mean and standard deviation. This approach provides a robust and objective measure of model performance.

## 3. RESULTS

### 3.1. Outlier Detection

To assess the impact of outlier detection on model performance, the IQR and Hampel filters were compared using RF as the baseline classifier. Table 1 summarizes the evaluation metrics for each method. The Hampel filter achieved slightly higher accuracy (85.42% vs. 85.03%) and F1-score (84.82% vs. 84.42%), while retaining more data (drop rate: 6.64% compared to 15.45% for IQR). These results demonstrate that the Hampel filter is more effective at identifying outliers while preserving useful data for subsequent analysis.

Table 1. Comparison of Outlier Detection Performance Using Random Forest

| Metric | IQR | Hampel |
|---|---|---|
| Accuracy (%) | 85.03 | 85.42 |
| Precision (%) | 84.52 | 84.81 |
| Recall (%) | 85.03 | 85.42 |
| F1-Score (%) | 84.42 | 84.82 |
| Drop Rate (%) | 15.45 | 6.64 |

### 3.2. Traditional Filtering Results

Traditional filtering methods were applied uniformly to the sensor data as a baseline for comparison. Table 2 presents the accuracy achieved by each filtering method across multiple machine learning models. The classification task focused on six behavior categories (Eating, Lying, Walking, Standing, and Interacting), with Drinking and Unknown excluded due to insufficient data.

Table 2. Accuracy of Traditional Filtering Methods Across Models (%)

| Method | RF | XGB | KNN | GBM | DT | Linear SVC | NB | Average Accuracy |
|---|---|---|---|---|---|---|---|---|
| Wavelet | **91.58** | **89.87** | **85.11** | **78.88** | **82.95** | 69.79 | 66.91 | **80.72** |
| Raw data | 84.69 | 83.14 | 81.96 | 78.65 | 76.33 | 73.40 | **80.95** | 79.87 |
| TVD | 84.48 | 83.06 | 82.01 | 78.45 | 76.39 | **73.68** | 67.35 | 77.91 |
| Median | 84.40 | 82.81 | 81.79 | 78.10 | 76.29 | 72.96 | 65.95 | 77.47 |
| HPF | 82.19 | 80.68 | 78.75 | 77.66 | 74.30 | 72.13 | 65.96 | 75.95 |
| Savitzky Golay | 82.96 | 81.09 | 79.83 | 76.08 | 73.51 | 69.15 | 62.64 | 75.03 |
| LPF | 82.66 | 81.14 | 78.81 | 76.04 | 73.44 | 68.24 | 61.07 | 74.48 |

The results indicate that wavelet is the only filtering method that improved classification performance relative to the raw data baseline (average accuracy of 80.72% versus 79.87%). In particular, dynamic models such as RF and XGB showed significant gains (91.58% and 89.87% accuracy, respectively) when using the wavelet method. In contrast, other methods, such as TVD (77.91%), median filtering (77.47%), and LPF (74.48%), failed to reach the baseline, suggesting that these techniques may overly smooth or distort key signal features.

Table 3 summarizes the average precision, recall, and F1-score achieved by each filtering method across all classification models. As shown, the wavelet method attained the highest average precision (81.55%), recall (82.55%), and F1-score (81.76%), outperforming the raw data baseline (precision: 77.14%, recall: 78.48%, F1-Score: 77.50%). The other filtering methods yielded intermediate to lower performance, with HPF, Savitzky Golay, and LPF yielding the lowest scores.

Table 3. Average Performance Metrics of Filtering Methods (%)

| Method | Precision | Recall | F1-Score |
|---|---|---|---|
| Wavelet | 81.55 | 82.55 | 81.76 |
| Raw data | 77.14 | 78.48 | 77.50 |
| TVD | 77.29 | 78.62 | 77.65 |
| Median Filter | 76.87 | 78.22 | 77.22 |
| HPF | 74.75 | 76.59 | 75.14 |
| Savitzky Golay | 74.90 | 76.01 | 75.09 |
| LPF | 74.39 | 75.52 | 74.51 |

These findings show that wavelet not only improves overall accuracy, but also achieves a better balance between precision and recall, as evidenced by its superior F1-score. This indicates that preserving high-frequency signal components is essential for accurately capturing dynamic pig behaviors, whereas aggressive smoothing may remove critical information. Such observations highlight the limitations of uniform filtering strategies and motivate the need for more tailored approaches. In the following section, we explore behavior-specific filtering methods designed to address the diverse characteristics of pig behavior data.

### 3.3. Behavior-Specific Filtering Results

Table 4 shows the performance of different behavior-specific filtering combinations across several models. The results clearly indicate that behavior-specific filtering improves classification performance compared to both single filtering methods and the raw data baseline.

Table 4. Performance of Behavior-Specific Filtering Combinations Across Models (%)

| Filtering Combination | RF | XGB | KNN | GBM | DT | Linear SVC | NB | Average Accuracy |
|---|---|---|---|---|---|---|---|---|
| Wavelet + Median | 93.87 | 94.71 | 90.70 | 90.18 | 87.09 | **84.39** | **77.43** | **88.33** |
| Wavelet + LPF | **94.11** | **94.73** | **91.03** | **90.28** | **89.84** | 83.27 | 72.36 | 87.94 |
| TVD + LPF | 88.91 | 88.45 | 87.05 | 84.57 | 79.61 | 80.76 | 75.24 | 83.51 |
| TVD + Median | 85.13 | 83.93 | 82.25 | 79.20 | 73.51 | 73.86 | 66.66 | 77.79 |

The results show that behavior-specific filtering markedly enhanced classification performance compared to both single filtering methods and the raw data baseline. In particular, the wavelet based combinations, wavelet + median and wavelet + LPF, yielded the highest average accuracies, with wavelet + median slightly outperforming its counterpart (88.33% versus 87.94%). These results are especially significant when contrasted with the standalone wavelet method, which achieved only 80.72% accuracy, underscoring the advantage of tailoring the filtering approach to distinct behavior types.

Moreover, the performance of the TVD-based combinations was notably lower, with average accuracies of 83.51% for TVD + LPF and 77.79% for TVD + median. This suggests that, although TVD can effectively reduce noise, it can also overly smooth the signal, thereby erasing critical features necessary for capturing dynamic behaviors.

The benefits of the wavelet-based combinations are further highlighted in dynamic models such as RF and XGB. For example, RF achieved up to 94.11% accuracy, and XGB reached 94.73% accuracy with these methods, indicating that preserving high-frequency signal components is vital for accurately recognizing dynamic pig behaviors.

Figure 2 illustrates the progression of classification performance, starting from raw data and moving through single filtering methods to the superior performance observed with behaviour specific filtering combinations. This progression clearly shows that while single filtering can improve performance, the highest gains are achieved when filters are tailored to the characteristics of active and inactive behaviors. Further supporting these findings, Figure 3 presents the confusion matrix for the XGB model using the wavelet + LPF combination. The matrix shows high classification accuracy for behaviors such as eating and laying, along with minimal misclassifications among similar behavior classes, further underscoring the effectiveness of the behavior-specific filtering approach in reducing errors.

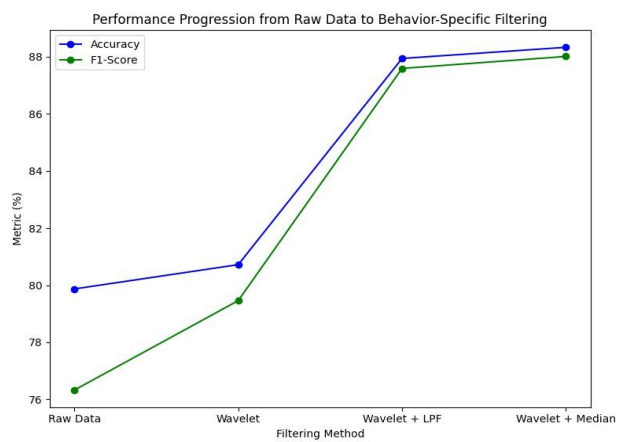

Figure 2. Progression of Accuracy and F1-Score from Raw Data to Behavior-Specific Filtering Combinations.

Overall, these results confirm that adapting filtering strategies to the unique attributes of different behavior types—especially by combining wavelet-based methods with noise reduction techniques—can significantly improve classification accuracy and reduce misclassification.

### 3.4. 10-Fold Cross-Validation Results

The 10-fold cross-validation results for the XGB model using the wavelet + LPF filtering method demonstrated a high and stable performance. On average, the model achieved an accuracy of 95.03%, with a precision of 91.90%, recall of 91.29%, and an F1-score of 91.57%. The low standard deviations, ranging from 0.16% to 0.39%, indicate that the model's performance is consistent across different folds. Figure 3a further illustrates these results, showing that the behavior-specific filtering approach reliably enhances the model's classification performance.

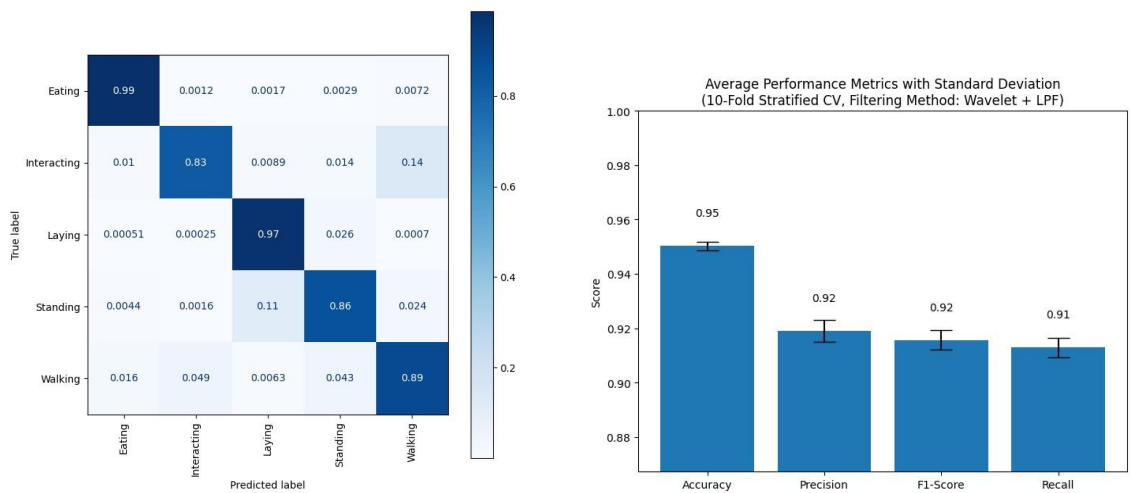

(a) XGBoost Confusion Matrix for Performance Metrics for XGBoost.

(b) 10-Fold Cross-Validation Wavelet + LPF Filtering.

Figure 3. (a) XGBoost Confusion Matrix from Wavelet + LPF Filtering and (b) Cross-Validation Performance Metrics illustrating model accuracy, precision, recall, and F1-Score.

## 4. DISCUSSION AND CONCLUSION

The results of this study demonstrate that behavior-specific filtering methods markedly improve the classification of pig behaviors compared to traditional, one-size-fits-all filtering approaches and raw data baselines. In particular, combinations such as wavelet + median filtering achieved the highest average accuracy (88.33%), substantially outperforming standalone wavelet denoising (80.72%) and the raw data (79.87%). These findings underscore the importance of tailoring filtering techniques to the unique signal characteristics of active and inactive behaviors.

Active behaviors, which exhibit high-frequency dynamics, benefit from the feature-preserving capabilities of wavelet denoising. In contrast, inactive behaviors require filters, such as LPF and median filtering, that emphasize noise suppression and signal stability. The superior performance of behavior-specific combinations (e.g., wavelet + LPF and wavelet + median) across various machine learning models, including dynamic models such as RF and XGB, highlights the effectiveness of this targeted approach in preserving critical signal features while reducing noise.

Beyond the improved classification performance, the practical implications for PLF are significant. Enhanced behavior classification facilitates early detection of health and welfare issues, supporting the development of scalable, automated monitoring systems. Additionally, the superior performance of the Hampel filter for outlier detection, resulting in only 6.64% data loss, demonstrates the value of robust preprocessing techniques in improving overall data quality.

Despite these promising results, several limitations warrant further investigation. The current study relied on manual categorization of behaviors, which may introduce bias. Future work should explore automated behavior classification to dynamically refine filtering strategies. Furthermore, the dataset, while sufficient for a proof-of-concept study, was limited in scale and scope, focusing solely on pigs. Expanding the dataset to include larger samples and diverse livestock species would enhance the generalizability of these findings. Finally, evaluating these methods in real-time settings is essential to address computational efficiency and practical deployment challenges in dynamic farm environments.

In conclusion, this study has shown that behavior-specific filtering represents a significant advancement in livestock behavior monitoring. By improving data quality and classification accuracy, these tailored methods pave the way for more effective and automated solutions in PLF, ultimately contributing to better animal health and welfare.

96Computer Science & Information Technology (CS & IT)

Computer Science & Information Technology (CS & IT)  97

## AUTHORS

**Zhen Zhang** received his B.S. in Computer Engineering (Machine Learning) from Virginia Tech in May 2024 and is currently pursuing his M.S. in Computer Engineering (Software Systems)at Virginia Tech. His academic research focuses on machine learning, computer vision, and precision livestock farming, with an emphasis on developing data-driven, robust preprocessing and filtering techniques for real-world applications.

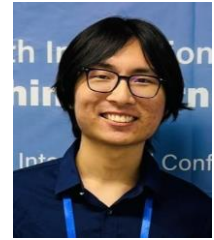

**Dong Sam Ha** received his B.S. degree from Seoul National University, Korea, and his M.S. and Ph.D. degrees from the University of Iowa. Since Fall 1986, he has been a faculty member in the Department of Electrical and Computer Engineering at Virginia Tech. His expertise lies in low power circuit and system design, focusing on analog and RF (Radio Frequency). Recently, he and his students have worked on wireless sensors for animal monitoring, RF-driven neural networks, and energy harvesting for biosensing and IoT applications. He is a Fellow of the IEEE.

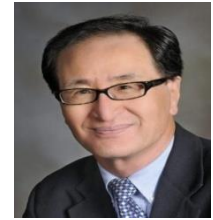

**Gota Morota** is an Associate Professor of Biometry in the Department of Agricultural and Environmental Biology at the University of Tokyo. His research focuses on quantitative genetics and phenomics with applications in animal and plant breeding. He received his Ph.D. from the University of Wisconsin-Madison in 2014

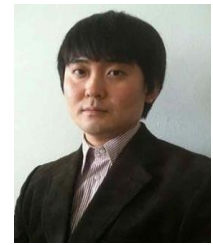

**Sook Shin\*** is a Collegiate Assistant Professor in the Department of Electrical and Computer Engineering at Virginia Tech. Her expertise is in bioinformatics, with a focus on gene and disease data analysis. Recently, her research has expanded to AI-driven precision livestock management, focusing on automated behavior analysis and predictive modeling for weight and body condition. She earned her Ph.D. from Virginia Tech in 2012.

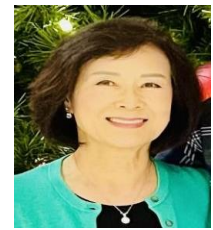